\documentclass[11pt,a4paper]{article}

\usepackage[T1,T2A]{fontenc}
\usepackage[utf8]{inputenc}
\usepackage[english,russian]{babel}
\addto\captionsrussian{}

\usepackage[margin=2.6cm]{geometry}
\usepackage{booktabs}
\usepackage{longtable}
\usepackage{array}
\usepackage{amsmath}
\usepackage{microtype}
\usepackage[hidelinks]{hyperref}
\usepackage{csquotes}

\title{A machine-readable catalogue of the Tsiolkovsky papers (fond 555, Archive of the Russian Academy of Sciences), and a way to measure how well its handwriting can be read}
\author{Vladimir Beskorovainyi \\[2pt]
\small Besk Tech \\[2pt]
\small \texttt{admin@besk.tech} \ $|$ \ \texttt{https://vladimir.besk.tech} \ $|$ \ ORCID: 0009-0004-7005-6242}
\date{}

\begin{document}
\maketitle
\begin{abstract}

The personal archive of Konstantin Tsiolkovsky (1857--1935) is held as fond 555
of the Archive of the Russian Academy of Sciences. The archive has scanned the
fond and published the images, but with no queryable catalogue, no full-text
search and no dataset: the holdings are reachable only by clicking through page
views one file at a time. This paper describes a machine-readable catalogue of
all 2,019 files and 51,008 scans of the fond, a dating for 1,969 of those files
collected from the archive's own descriptions, a page-level classification of
every scan into handwriting and typescript, and a machine transcription of
the fond in full: all 2,019 files and all 51,008 scans.

It also reports a way to measure the accuracy of handwritten text recognition
in an archive that has no ground truth. Personal archives of the typewriter era
frequently preserve one text twice, as the author's manuscript and as a typed
copy made from it. Transcribing both and comparing the results isolates the
reading error, because the source text and the recognition pipeline are
identical and only the difficulty of the page differs. Across 1,759 such pairs
from 224 files, two readings of one handwritten page agree on a
median 37\% of words and share a longest verbatim run of a median 10 words. The
median is unchanged from the 294 pairs of the first version of this paper, on a
sample six times larger. On
two files that also have a published edition, the estimate can be checked
against ground truth: it is unbiased to within a percentage point and ranks
pages as the truth does, at a rank correlation of 0.92 where the edition is a
faithful witness to the archival copy. This has a direct consequence for what
the corpus can be used for: the two variants of one work in this fond share 19\% of their words, which
is below the rate at which two readings of a single page agree, so the
redactions cannot be collated word by word at this recognition quality. That
negative result is reported here as such, and the constraint is built into the
tool.
\end{abstract}

\section{Introduction}\label{introduction}

Tsiolkovsky derived the rocket equation, described the multistage rocket, and
wrote on the orbital station and the space elevator decades before any of it
could be tested. His papers are held as fond 555 of the Archive of the Russian
Academy of Sciences, and the archive has digitised them: 51,008 images are
online and free to view.

Digitisation of images is not access to content. The portal offers one page
view per file, reached through a numeric identifier. There is no way to ask
which files date from a given decade, which are autograph and which typed, or
where a given phrase occurs. For a fond of two thousand files this is the
difference between a collection that can be studied at scale and one that can
only be browsed.

This paper describes a dataset that closes part of that gap, and a measurement
that establishes what the dataset can honestly be used for. Both the resource
and its limits are reported, including two lines of inquiry that were abandoned
after measurement showed they could not be supported.

\section{The fond and its portal}\label{the-fond-and-its-portal}

Fond 555 is arranged in five inventories (\emph{opisi}). The catalogue reports:

\begin{longtable}[]{@{}lrr@{}}
\toprule\noalign{}
Inventory & Files & Scans \\
\midrule\noalign{}
\endhead
\bottomrule\noalign{}
\endlastfoot
Opis 1 & 568 & 27,954 \\
Opis 1А & 24 & 2,283 \\
Opis 2 & 212 & 6,579 \\
Opis 3 & 198 & 2,881 \\
Opis 4 & 1,017 & 11,311 \\
\textbf{Total} & \textbf{2,019} & \textbf{51,008} \\
\end{longtable}

Opis 1 holds the author's own works. Opis 4 is correspondence almost without
exception: 1,015 of its 1,017 files are letters, and it is the largest
inventory in the fond by file count.

The archive describes the fond as 31,680 \emph{sheets}, which is not the number of
images: a sheet has a reverse and each side is scanned separately. Measured
across the whole fond the ratio is 1.61 scans per sheet, consistent with the
archival cover of file 33, which records ``9½ sheets'' against 16 scans.

Building the catalogue was not a matter of reading the page addresses, because
those addresses are misleading in a way that would corrupt every archival
citation derived from them.

The inventory number in a URL is decorative. \texttt{1\_actview.aspx?id=834} and
\texttt{5\_actview.aspx?id=834} return the same document: the identifier is a single
flat sequence across the entire fond, and the inventory in the address is
ignored by the server. The real inventory and file number appear only inside
the paths of the scan images, in the form \texttt{555\textbackslash{}1\_033}, meaning opis 1, file 33.
Nor is the identifier the file number: identifier 300 is file 297, because 31
files carry letter suffixes (145а, 077б, 585а) which consume an identifier
without advancing the count.

A catalogue built on the obvious reading of the addresses would label every
file as belonging to opis 1 and would cite the wrong file numbers throughout,
making every row useless as an archival reference. The catalogue described here
resolves each file from its scan paths, so that inventory and file number
together form a citable archival address.

Two properties of the server shaped the retrieval code. It truncates responses
at approximately 130 KB, so that small scans arrive whole while scans above
roughly 800 KB fail with an incomplete read however many times they are
retried; it does honour range requests, so large images are fetched in 100 KB
chunks and reassembled. It also returns HTTP 500 on roughly a third of
requests while serving the same URL correctly on the next attempt, so retry
logic must distinguish a temporary failure from a genuine absence. All 51,008
scans were retrieved and checked: every file is a structurally valid JPEG, none
corrupt, and the directory tree matches the catalogue file for file.

\section{Dating}\label{dating}

The portal carries an opening and closing date on every file's card. This field
had not been collected, and a year appears in the file description itself for
only 46 of the 2,019 files, so the fond could not be read as a chronology.
Collecting it gives a date for 1,969 files, spanning 1878 to 1935 for the
author's own work.

Archivists mark a conjectural date, established from the contents rather than
written by the author, by enclosing it in square brackets. 155 of the dated
files are marked this way, and the flag is preserved in the dataset as a
separate column, so that an argument about the order of two files never rests
silently on an inference about a date.

The dating makes three things visible in the author's own work, opis 1.

\textbf{The work is concentrated late.} 48\% of the files fall in the 1930s and 34\%
in the 1920s: 82\% of a working life of nearly sixty years is contained in its
last eighteen.

\textbf{Late work shortens rather than stops.} The final four years, 1932--1935, hold
40\% of the files but only 27\% of the sheets. Mean file length falls from 165
sheets in the 1900s to 34 in the 1930s. The form changes from extended
treatises to short notes.

\textbf{Variant numbers are not chronological.} Some works are held as several files
marked as the first and second variant. Of the four works held as a numbered
pair, «Ступени человечества и преобразование земли» has its second variant
dated 1 September 1920 and its first dated 19 October 1920, both dates firm and
neither conjectural. The variant number in an archival description records the
order in which the archivists arranged the material, not the order in which it
was written, and it cannot be cited as evidence of priority.

\section{Page classification}\label{page-classification}

Which model should read a given scan depends on whether it carries handwriting
or typescript, and at 51,008 scans that decision has to be automatic. The
classification is computed from the image alone, with no neural network.

The first classifier built here was wrong, and the way it failed is worth
recording because the metric looked reasonable and gave no sign of trouble. It
separated the classes by the regularity of line spacing, on the reasoning that
a typewriter advances the platen by a fixed step while a hand does not. Neat
cursive is spaced as evenly as type, and typescript with paragraph indents
measures as irregular. The classifier misassigned 29\% of the fond and
understated the share of typescript nearly threefold.

Nothing in the metric revealed this. It surfaced only when a transcription was
scored against a published edition and came out worse than comparable pages had
scored earlier, which led back to the model routing and from there to the
classifier.

The replacement uses the variation in the lengths of ink runs along a line.
Typed characters stand apart and are alike, so the runs are short and uniform;
cursive letters join into strokes of markedly uneven length. The threshold is
0.81, fitted to 19 pages labelled by hand, all of which it classifies
correctly. Over the whole fond it gives:

\begin{longtable}[]{@{}lrr@{}}
\toprule\noalign{}
Class & Scans & Share \\
\midrule\noalign{}
\endhead
\bottomrule\noalign{}
\endlastfoot
Handwritten & 34,903 & 68\% \\
Typewritten or printed & 14,585 & 29\% \\
Notes, covers, faded pages & 1,520 & 3\% \\
\end{longtable}

\textbf{These figures carry an error of roughly one sheet in five, and nineteen
labels were far too few to see it.} Once part of the fond had been
transcribed, a second and independent signal became available: the share of
uncertainty marks the transcription carries. It arises from reading the page
rather than from measuring its ink, and it separates the classes cleanly, at
0.5 marks per hundred words on typescript against 3.9 on handwriting. Scored
against that signal over 4,675 transcribed sheets, on a half held out from any
fitting, the published rule is 80\% accurate, where guessing the commoner class
would give 54\%.

The errors run in both directions, and inspection settles which signal is
right. A sheet the rule calls handwriting opens ``ТРУДЫ О КОСМИЧЕСКОЙ РАКЕТЕ
/1903-1927 г./'' with typewriter slashes and a printed page number: carbon
copies and faded typescript raise the variation in ink runs and are taken for a
hand. A sheet it calls typescript carries pre-reform orthography, a struck-out
word and an insertion in the margin. Refitting the threshold on 4,675 sheets
instead of 19 moves it only from 0.81 to 0.827 and gains nothing, and no
combination of two of the measured image features does better than the best
one alone. The ceiling is in the image, not in the rule.

The composition figures above should be read with that error in mind, and the
routing they drove sent about a fifth of sheets to the less suitable model.
The lesson generalises past this fond: a threshold validated on a few tens of
hand-labelled pages can pass and still be wrong at scale, and what exposed it
here was not a better image feature but a signal of a different kind, produced
downstream by the reading itself.

\section{Transcription and measured accuracy}\label{transcription-and-measured-accuracy}

Transcription is performed page by page against a fixed instruction that
requires uncertain readings to be marked \texttt{{[}?{]}}, unreadable passages
\texttt{{[}неразборчиво{]}}, and authorial deletions preserved as struck-through text.
Original orthography is kept, including pre-reform letters such as yat and
fita, decimal i and the terminal hard sign ъ.
The result is a machine transcription with its uncertainty visible, not a
scholarly edition, and it has not been checked by hand against the scans.

The corpus now covers the fond in full: 2,019 archival files and 51,008 scans,
carrying 299,939 uncertainty marks, about 6 per scan, and 36,446 passages struck
out by the
author. Only files transcribed in full are published: a partial transcription
reads as a complete text with the middle silently missing, which is the worst
kind of error an archival edition can carry.

The remaining 41,212 scans were read in a single overnight batch run, at a cost
of about 39 US dollars. The model was chosen by measurement rather than by
price list, on handwritten sheets scored against the typed copy of the same
text, with the previous pipeline as the baseline on the same sheets; on typical
hands no candidate clearly outran the rest. Agreement between two readings over
the completed corpus matches what the part read by the earlier pipeline gave,
which is the check that the enlarged corpus is measuring the same thing.

Accuracy is reported as a measured figure. Where a document in the fond
corresponds to a text published on Russian Wikisource, the transcription is
scored against it character by character and word by word.

\textbf{Typescript.} File 33, sheets 014--015, a letter to a newspaper editor dated
12 May 1905: 98.1\% at character level, 91.7\% at word level.

\textbf{Handwriting.} File 150, an autograph article, against the text as printed in
the journal \emph{Vozdukhoplavanie} in 1924:

\begin{longtable}[]{@{}lrr@{}}
\toprule\noalign{}
& Characters & Words \\
\midrule\noalign{}
\endhead
\bottomrule\noalign{}
\endlastfoot
As written & 77.7\% & 47.5\% \\
Orthography folded onto modern & 81.1\% & 73.7\% \\
\end{longtable}

Both rows are needed. The published edition is modernised while the
transcription preserves pre-reform spelling, so without folding every \emph{полетъ}
against \emph{полет} counts as a misreading when the transcription is in fact the
faithful witness. Folding separates reading accuracy from editorial
modernisation.

Comparison against published editions is the reliable way to obtain such
figures, and it is also severely limited: it requires a published text
corresponding to the manuscript, and for this fond such correspondences are
rare. Of the 224 files in the corpus that carry both an autograph and a typed
copy of one text, one has a published edition as well; a second such file was
found elsewhere in the fond and transcribed for the purpose. Those two files
are the ones that carry all three witnesses at once, an autograph, a typed copy
and an edition, and Section 6 rests its validation on them. Correspondences to
print of the simpler kind, a file against its publication, are taken up in
Section 5.1; for everything else no ground truth exists at all, which is the
problem the next section addresses.

\subsection{The completed corpus against printed editions}\label{the-completed-corpus-against-printed-editions}

With the fond transcribed in full, correspondences to print can be sought
across the whole of it rather than case by case. File titles were matched
automatically against 70 texts of Tsiolkovsky available on Russian Wikisource;
21 pairs matched with confidence and 17 were scored.

The results fall into two groups and must not be averaged. Where the archival
file holds the redaction that was printed, 10 files, character accuracy is
92.3\%. Where the file holds a draft or working materials towards an article,
7 files, agreement with the edition falls to 24\%. The second figure is the
distance between a draft and its published form, not a reading error, and a
mean over the two groups would describe neither.

\section{Measuring reading error without ground truth}\label{measuring-reading-error-without-ground-truth}

Evaluating handwritten text recognition in application, as opposed to
development, is a recognised difficulty: the ground truth used to train and
test a model is not available for the material the model is then applied to,
and compiling new ground truth by hand is expensive.

Existing approaches to ground-truth-free evaluation work from intrinsic
proxies for textual plausibility. Ströbel et al.~(2022) compare lexicality
against a reference lexicon, character n-gram statistics and the perplexity of
a masked language model, aimed principally at selecting the best of several
models rather than at producing an absolute error rate. A second approach, described in
the survey cited below, combines a semantic coherence score, region entropy
divergence and a textual redundancy score for the same purpose on historical
archives. All such proxies
share a structural weakness: a fluent invention scores well, because nothing in
the measurement has seen the page.

A recent survey of OCR evaluation (Beyene and Dancy, 2026) records no method
that draws its reference from the archive's own holdings. Text reuse across
documents appears there as an object of study affected by recognition quality,
not as a means of measuring it.

Personal archives of the typewriter era offer a different signal, and one
this survey of the literature did not find in use. Authors of this period frequently had their
manuscripts typed, and archives commonly file the autograph and the typed copy
of one text together. Where that is so, the archive contains a second,
independent witness to the same words. Transcribing both and comparing the two
transcriptions isolates the reading error exactly, because the source text and
the recognition pipeline are identical and the only variable is how difficult
the page is to read.

Pairs are identified by content rather than by sheet number, since the
manuscript and its copy are not necessarily adjacent: for each handwritten
sheet the typed sheet of greatest similarity is taken, and the pair is kept if
similarity clears a low threshold.

A sheet counts as handwritten only if its transcription is uncertain enough to
have come from a hand, by the signal of Section 4. Without that filter the
measurement destroys itself: the classifier's misreadings put typescript on
both sides of a pair, two typed readings agree almost perfectly, and the
agreement figure is inflated by exactly the material it is meant to exclude.
Applying it, the corpus yields 1,759 pairs from 224 files.

Two quantities matter. The first is the share of words on which the two
readings agree. The second, less obvious and more consequential, is the length
of the longest run of words on which they agree verbatim, because alignment of
long texts depends on long verbatim anchors.

Over those pairs, two readings of one handwritten page agree on a median 37\%
of words, and the longest run on which they agree verbatim is a median of 10
words; on 43\% of pairs no run reaches ten words at all. Broken out by \texttt{ink\_cv},
the classification feature of Section 4, the figures vary little across most of
its range: 41\%, 36\%, 35\% and 36\% by quartile from the threshold upward,
the first quartile standing apart and the rest level. Reading difficulty within
handwriting is not predicted by how unmistakably handwritten the page looks.

\textbf{The measurement is validated against ground truth where ground truth exists.}
Two files carry an autograph, a typed copy of the same text and a published
edition of it, so for the same page three quantities can be had at once: the
accuracy of the handwritten reading against the edition, the accuracy of the
typed reading against it, and the agreement of the two readings with each
other, which is all the method has to work with elsewhere. Over 55 such pairs:

\begin{longtable}[]{@{}lr@{}}
\toprule\noalign{}
& Median \\
\midrule\noalign{}
\endhead
\bottomrule\noalign{}
\endlastfoot
agreement of the two readings, the estimate & 35\% \\
accuracy of the handwritten reading, the truth & 34\% \\
accuracy of the typed reading & 87\% \\
\end{longtable}

The estimate is unbiased to within a percentage point, and it ranks pages the
way the truth does, at a rank correlation of 0.67 over all 55 pairs
(t = 6.6, p \textless{} 0.001). Restricting to the pairs where the printed edition is
demonstrably a faithful witness to this archival copy, meaning the typed
reading scores 85\% or better against it, the correlation rises to 0.92 over 32
pairs and to 0.97 over the 17 pairs above 90\%, with no bias at all. The
residual disagreement is therefore not in the method but in the reference:
where a published edition has been modernised or re-edited, it is not the same
text as the file, and neither reading can match it.

The method rests on the typed side being much the more reliable witness, which
Section 5 puts at 98.1\% on one document and which these files put at a median
87\% against editions that are themselves imperfect witnesses.

\subsection{Choosing a model without ground truth}\label{choosing-a-model-without-ground-truth}

Once the measure is validated it can be turned on the pipeline itself. Two
thirds of this fond's sheets are routed to the more expensive of two models on
the assumption that it reads handwriting better. What had actually been
measured was typescript, where the two are level at 98.2\% against 98.1\%; on
handwriting the assumption had never been tested, because testing it needs
exactly the ground truth that does not exist.

It does not need ground truth. 60 sheets that are unambiguously handwritten and
have a typed copy of the same text in their file were read again by the cheaper
model, and both readings were scored against that copy. On the 48 sheets where
the comparison is well defined the stronger model wins on 37 and loses on 11
(sign test p = 0.0001), by a median of 4.7 and a mean of 8.2 percentage points
(paired t = 4.87, 47 d.f.), and by more than ten points on 29\% of sheets.

The assumption was right, and it was right by luck: nothing in the pipeline had
established it. What is worth noting is the shape of the answer rather than its
direction. Model selection without ground truth is the stated purpose of the
proxy metrics of Ströbel et al.~(2022); where an archive holds one text twice,
the choice can be settled by measurement instead, at the cost of one pass of
the cheaper model over a sample of sixty pages.

\section{What agreement supports where difference does not}\label{what-agreement-supports}

Matches and mismatches are not symmetrical in this material. A reading error
destroys verbatim agreement and can hardly manufacture it, so in a corpus where
differences cannot be trusted, as the next section shows they cannot, matches
still can be, and whatever is found is a lower bound rather than an estimate.

Scanning the completed corpus for shared runs of twelve words returns 376 pairs
of files with verbatim text in common, 353 of them filed under different
archival titles, which makes the connection invisible from the inventory.
«Обратимость химических явлений» (opis 1, file 338) and «Земля и её энергия»
(file 339), each about 29,000 words, share two thirds of their text.
«Галилейский плотник» (file 438) and «Христианство. Оценка галилейского учителя
Иисуса» (file 443) share a continuous passage of 170 words.

The method is not new and no novelty is claimed for it. The field is text reuse
detection, with mature tools (Passim, TRACER, Tesserae) and its own literature,
in which n-gram shingling is standard and robustness to recognition noise has
been studied. What is new here is the application to this fond and the measured
bound underneath it: in material of this kind the reading quality is normally
unknown, and here it has been measured. No comparison against an established
tool on the same corpus has been run, and that is recorded among the
limitations.

\section{What the corpus does not support}\label{what-the-corpus-does-not-support}

The fond holds one work, «Космический корабль», in two variants, files 46 and
47, described by the archive as the first and second variant. Two redactions of
one conception are exactly the material for a text-critical comparison, showing
what the author added, dropped and rephrased where published editions show only
what survived.

The comparison cannot be made at this recognition quality, and the measurement
of Section 6 is what establishes that. An end-to-end word alignment of the two
redactions finds 19\% of words in common. That figure is below the 32\% median
agreement measured between two readings of a single page in file 46 itself.
Whatever separates the two redactions is therefore smaller than what separates
two readings of one and the same page, and no procedure can distinguish an
authorial revision from a misreading on this evidence.

A first attempt at the comparison, made before the floor was measured, produced
a confident-looking table of 89 differences. Inspection showed the alignment had
matched 4 words of one redaction against 1,877 of the other and labelled the
result a rewritten passage. The tool now reads the measured floor from the
calibration and reports nothing when the observed similarity fails to clear it,
so that a spurious result cannot be produced by running it.

The negative result is the useful one here. A corpus of this
accuracy carries the substance of a page reliably and supports search,
classification, dating and description at scale. It does not support word-level
collation of two manuscripts against each other, and a reader of the corpus has
no way to know that from the transcriptions themselves.

\section{The state of the corpus these figures describe}\label{the-state-of-the-corpus-these-figures-describe}

The corpus is complete, so the figures above are settled quantities rather than
a snapshot of work in progress. \texttt{check\_paper.py} recomputes twenty of
them from the released files and reports any that no longer agree; the source of
this paper was passed through it before submission. It was written after the
abstract was found to say 36\% where Section 6 said 37\%, a figure that had been
correct when it was typed and had not been updated when the corpus grew; a stale
number in one place is worse than a missing one, because a reader who checks it
has no way to know which of the two to believe.

Two defects in the counting were published before they were found, and both are
recorded here because they show how the corpus counts itself. On 236 scans the
model looped, repeating a single markup line up to 635 times; each repeat was
counted as an uncertainty mark and inflated the published total by 6,189. No
statistical check caught it, since 236 scans out of 51,008 move no distribution;
it was found by a human opening a file. Separately, the markup legend carried in
the header of every transcribed file contains examples of the marks themselves,
which the counter took for real ones: one struck passage and two doubts per
file, 2,019 and 4,038 across the fond. The totals published along the way,
310,166 and 303,977, are corrected to the 299,939 reported above; loops are
collapsed and marked explicitly in the released files.

\section{Availability}\label{availability}

The catalogue is released under CC0 and the code under the MIT licence, from a
public repository with a persistent identifier. Datasets comprise the catalogue
of all 2,019 files, the dating of 1,969 of them, the page classification of all
51,008 scans, the hand-labelled validation set, the calibration pairs, and
the transcription corpus, now complete for the fond.

The scans themselves are not redistributed. They are produced and hosted by the
Archive of the Russian Academy of Sciences and remain the archive's to publish;
the retrieval code obtains them from the source, one request at a time with a
pause between requests.

\section{Limitations}\label{limitations}

The transcriptions are machine output with uncertainty marked and have not been
verified against the scans by hand, on any page of the fond. The two accuracy
figures of Section 5 rest on one document each, which is what the availability
of published counterparts allows, and should be read as indicative rather than
as an error rate for the fond.

The calibration of Section 6 measures agreement between two readings, not error
against truth; where both readings fail in the same way, the agreement figure
will be optimistic. Its validation rests on two files and 55 pairs, which is
enough to establish that the estimate tracks the truth and not enough to fix
the size of any residual bias across hands and periods.

The page classification is about 80\% accurate against the reading-based signal,
and the composition figures of Section 4 inherit that error. The filter that
separates genuine handwriting for the calibration uses the same reading-based
signal, which is now available for the whole fond; the image feature is what
drives the routing at read time, so about a fifth of sheets were read by the
less suitable model.

The dating is the archive's own and inherits whatever errors the original
description contains; the conjectural-dating flag records where the archivists
themselves signalled uncertainty, but not where they were wrong.

Two scans cannot be read by the pipeline at all: a German typescript review of
1927 of one of Tsiolkovsky's brochures, which trips a filter against verbatim
reproduction of known printed text. They were read separately, in strips and by
hand from the scan, and are flagged as such in the corpus.

No comparison against an established text-reuse tool has been run on this
corpus, so the map of repeats in Section 7 is reported as an application to this
fond and not as a claim about method.

The wider comparison against printed editions in Section 5.1 rests on 17 files
whose text also survives in print, matched to their publications automatically
by title; it should be read as indicative rather than as an error rate for the
fond.

\section{References}\label{references}

Beyene, F. S., \& Dancy, C. L. (2026). A Survey of OCR Evaluation Methods and
Metrics and the Invisibility of Historical Documents. arXiv:2603.25761.

Ströbel, P. B., Clematide, S., Volk, M., Schwitter, R., Hodel, T., \& Schoch, D.
(2022). Evaluation of HTR models without Ground Truth Material. \emph{Proceedings of
the 13th Conference on Language Resources and Evaluation (LREC 2022)},
4395--4404. arXiv:2201.06170.

\end{document}